\documentclass[sigconf,nonacm]{acmart}

\renewcommand\footnotetextcopyrightpermission[1]{} 
\setcopyright{none}

\renewcommand\footnotetextcopyrightpermission[1]{} 
\usepackage[table,xcdraw]{xcolor}

\usepackage{svg}

\usepackage{multirow}
\usepackage[table]{xcolor}
\usepackage{colortbl}
\usepackage{graphicx}
\usepackage{algorithm}
\usepackage[noend]{algpseudocode}
\algrenewcommand\algorithmicindent{1.0em}%

\usepackage{enumitem}
\setlist[itemize]{leftmargin=*}

\usepackage{comment}
\usepackage{soul}

\newcommand{\ckblue}[1]{} 
\newcommand{\respblue}[1]{} 
\newcommand{\ignore}[1]{}

\usepackage{ulem}

\usepackage{booktabs}   
\usepackage{multirow}   
\usepackage[table]{xcolor} 
\usepackage{siunitx}   

\AtBeginDocument{%
  }

\begin{document}

\title{GPS-MTM: Capturing Pattern of Normalcy in \\GPS-Trajectories with self-supervised learning}


\author{Umang Garg}
\affiliation{%
  \institution{ECE Department}
  \institution{University of California}
  \city{Santa Barbara}
  \country{USA}
}
\email{umang@ucsb.edu}

\author{Bowen Zhang}
\affiliation{%
  \institution{ECE Department}
  \institution{University of California}
  \city{Santa Barbara}
  \country{USA}
}
\email{bowen68@ucsb.edu}

\author{Anantajit Subrahmanya}
\affiliation{%
  \institution{ECE Department}
  \institution{University of California}
  \city{Santa Barbara}
  \country{USA}
}
\email{anantajit@ucsb.edu}

\author{Chandrakanth Gudavalli}
\affiliation{%
  \institution{ECE Department}
  \institution{University of California}
  \city{Santa Barbara}
  \country{USA}
}
\email{chandrakanth@ucsb.edu}

\author{B. S. Manjunath}
\affiliation{%
  \institution{ECE Department}
  \institution{University of California}
  \city{Santa Barbara}
  \country{USA}
}
\email{manj@ucsb.edu}

\renewcommand{\shortauthors}{Garg et al.}

\newcommand{\uncited}{\textcolor{red}{\textbf{[cite]}}}





\begin{abstract}


Foundation models have driven remarkable progress in text, vision, and video understanding, and are now poised to unlock similar breakthroughs in trajectory modeling. We introduce the GPS-Masked Trajectory Transformer (GPS-MTM), a foundation model for large-scale mobility data that captures \textit{patterns of normalcy} in human movement. Unlike prior approaches that flatten trajectories into coordinate streams, GPS-MTM decomposes mobility into two complementary modalities: \textit{states} (point-of-interest categories) and \textit{actions} (agent transitions). Leveraging a bi-directional Transformer with a self-supervised masked modeling objective, the model reconstructs missing segments across modalities, enabling it to learn rich semantic correlations without manual labels. Across benchmark datasets, including Numosim-LA, Urban Anomalies, and Geolife, GPS-MTM consistently outperforms on downstream tasks such as trajectory infilling and next-stop prediction. Its advantages are most pronounced in dynamic tasks (inverse and forward dynamics), where contextual reasoning is critical. These results establish GPS-MTM as a robust foundation model for trajectory analytics, positioning mobility data as a first-class modality for large-scale representation learning. Code is released for further reference.\footnote{\url{https://github.com/umang-garg21/GPS-MTM}}

\end{abstract}



\received{TBD}
\received[revised]{TBD}
\received[accepted]{TBD}

\maketitle

\section{Introduction}

\begin{figure}[t]
 \centering
 \includegraphics[width=\columnwidth]{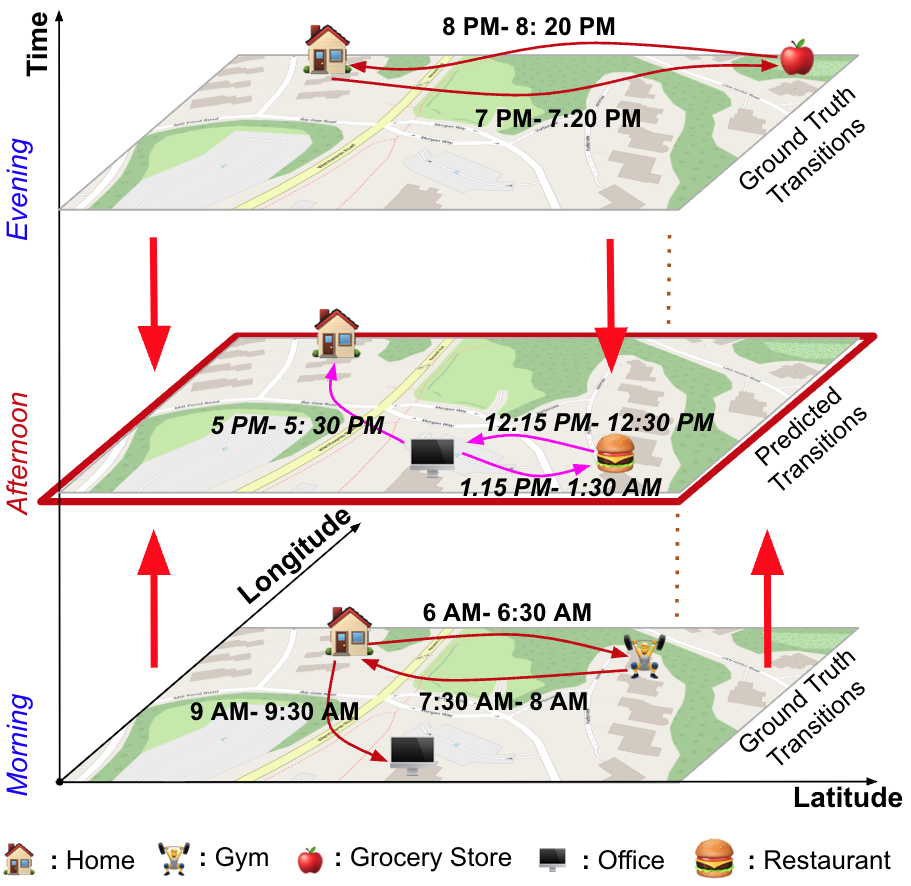}
 \caption{Multimodal representation of mobility trajectories. Daily activities are expressed as \textbf{states} (dwelling at specific locations) and \textbf{actions} (transitions between them). Our model learns to reconstruct missing components, aligning predicted transitions with ground truth movement patterns.}
 \label{fig:mtm_intro}
\end{figure}

The ubiquity of GPS-enabled devices has resulted in unprecedented amounts of trajectory data being collected every day. These trajectories carry rich semantic information about daily routines and collective movement patterns. Modeling this information is critical for applications in urban planning~\cite{TA2016161, mohammadi_urban_2016, bhat_household-level_2013}, public health~\cite{hast2019use, lai_measuring_2019}, policy~\cite{keenan_state_2025, zhang_effect_2021}, and personalized services, building upon a long history of GPS-based mobility research~\cite{draijer_global_2000, gonzalez_understanding_2008, schneider_unravelling_2013, ye_mining_2009, wu2019inferring}. However, the key challenge is that trajectory data is largely unlabeled; manual annotation is expensive, ambiguous, and does not scale. This motivates the development of \emph{self-supervised learning} (SSL) methods that can exploit large-scale GPS traces without costly supervision.

Recent progress in \emph{foundation models} has reshaped representation learning across text, vision, and speech modalities~\cite{devlin2019bert, radford2021clip}. Yet, trajectory modeling as a primary modality remains at an early stage. Early works~\cite{trajgpt, hsu_trajgpt_2024, deepam, liao_deep_2024, qin_transgte_2024, jin_urban-epr_nodate, lu_self-attention_2023, yang_getnext_2022, ying_semantic_2011, li_spabert_2022, liu_nextlocllm_2024, feng_agentmove_2025} adapted transformer architectures and other deep learning techniques to mobility data, demonstrating encouraging results. However, these methods typically flatten mobility into streams of coordinates or grid cells. Such representations can capture short-term transitions but often miss the semantic, hierarchical structure of daily life, struggling with geospatial generalization~\cite{tenzer_geospatial_2023}.

A parallel line of research has demonstrated the value of integrating multiple modalities during pretraining. For example, the MTM framework~\cite{mtm} shows that aligning heterogeneous signals yields richer representations. Motivated by this insight, we observe that mobility data itself contains inherent modalities: \textbf{states} and \textbf{actions}. Unlike externally paired data, these arise organically from trajectories. Modeling trajectories in this multi-modal form, as illustrated in Figure~\ref{fig:mtm_intro}, not only reflects how humans reason about activities but also enables powerful cross-modal prediction tasks, positioning mobility as a first-class modality for foundation-model-level representation learning.

In this work, we argue that trajectories can be naturally decomposed into two complementary modalities: \textbf{states}, which describe where an agent dwells (e.g., \textit{home, gym, work}), and \textbf{actions}, which capture semantics related to those states (e.g., \textit{arrival/departure times}). This state–action formulation provides a structured basis for self-supervision. We adopt a masked modeling strategy in which portions of one modality are hidden and reconstructed from the other. For example, as shown in Figure~\ref{fig:mtm_intro}, if we observe an agent's morning (6 AM--11 AM) routine of $\langle\text{\textit{home, gym, home, work}}\rangle$ and evening (6 PM--10 PM) activities of $\langle\text{\textit{home, grocery store, home}}\rangle$, the model can predict the most likely pattern for a missing afternoon (12 PM--5 PM) segment. It might infer, for instance, a commute to a \textit{restaurant} with arrival at 12:20 PM and a return to \textit{work} around 1:30 PM. Such cross-modality reasoning enables GPS-MTM to capture semantic patterns of daily life—\emph{patterns of normalcy}~\cite{zhang_reespot_2024, gudavalli_reeframe_2024}—in a scalable and label-free manner, with states identified via methods like clustering~\cite{narita_incremental_2018}.

This formulation grounds raw GPS traces in semantically meaningful constructs, captures long-range temporal dependencies, and enables strong generalization to diverse tasks such as trajectory infilling, next-stop prediction~\cite{zhang_individual_nodate}, and anomaly detection~\cite{liu_online_2020, wen_uncertainty-aware_2025, cruz_applying_2023, amiri_urban_2024}, with evaluation supported by specialized metrics~\cite{shimizu_geo-bleu_2022}. This perspective positions mobility data alongside text and vision as a modality where foundation models can unlock powerful representation learning.

\noindent
\textbf{Contributions.} Building on these insights, this paper makes three key contributions:
\begin{itemize}
\item A bi-directional Transformer framework with a multi-modal state–action representation for trajectory modeling.
\item A self-supervised masked modeling objective that captures semantic \emph{patterns of normalcy} without labeled data.
\item An augmented GeoLife dataset with 198 unique POI categories, adding to the ecosystem of open resources vital for mobility research~\cite{yabe_human_2023, yabe_metropolitan_2023, yabe_yjmob100k_2024, stanford_numosim_2024, kim_humonet_2024, noauthor_cruiseresearchgroupmassive-steps_2025, wang_libcity_2021, feng_vonfengdeepmove_2025, Krajzewicz2010, valhalla_engine, valhalla_mjolnir_datasources}.
\end{itemize}

Together, these contributions establish GPS-MTM as a foundation-model framework for mobility data and set the stage for our detailed description of the proposed method.
\section{Proposed Method: GPS-MTM}
\label{sec:methodology}

\subsection{Overview}
The widespread use of GPS tracking is often hampered by signal dropouts, resulting in incomplete trajectories. Our goal is to accurately reconstruct these missing segments by framing trajectory recovery as an \textbf{infilling task}, where the model predicts masked (missing) stop points conditioned on the observed ones~\cite{hsu_trajgpt_2024}.  

The proposed GPS-Masked Trajectory Transformer (GPS-MTM) addresses this challenge by combining two insights. First, human mobility is inherently \textbf{multi-modal}: a trajectory reflects not just locations, but also the \textbf{type of activity} (e.g., work, social) and its \textbf{temporal context} (e.g., arrival time, duration)~\cite{deepam}. Second, mobility patterns can be naturally decomposed into complementary \textbf{states} (dwelling at points of interest) and \textbf{actions} (transition details encoded as times of arrival, departure, and stay)~\cite{mtm}.  

\begin{figure}[t]
    \centering
    \includegraphics[width=1.0\columnwidth]{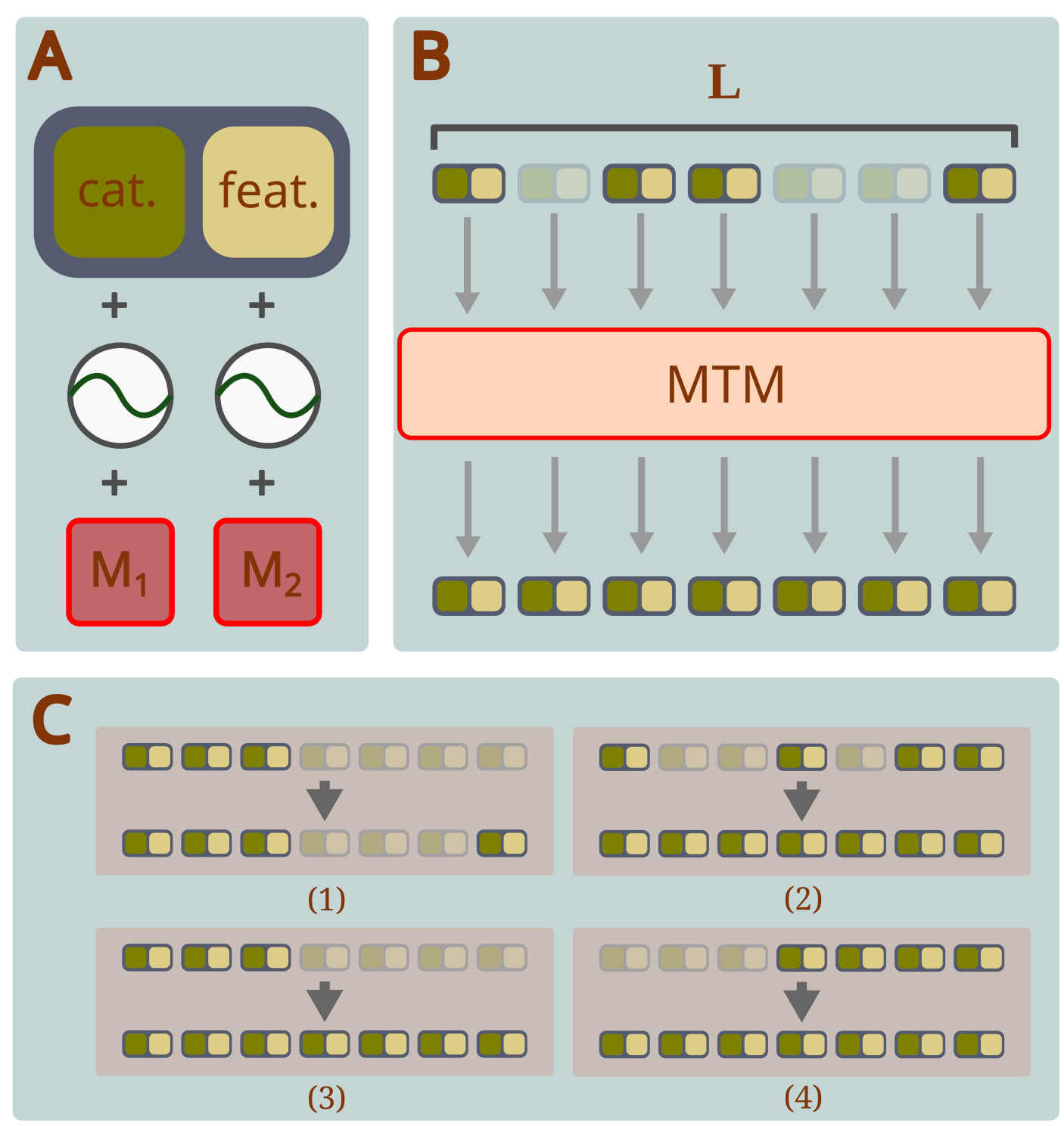}
    \caption{\textbf{Multi-modal trajectory representation and pre-training framework.} (A) Token structure with category and feature components, each enhanced with positional and modality embeddings. (B) Masked token reconstruction during pre-training, where a subset of L input tokens are randomly masked and the model learns to reconstruct the missing tokens from the remaining context. (C) Four downstream tasks used for evaluation: [1] goal prediction, [2] random masking, [3] forward dynamics, and [4] inverse dynamics.}
    \label{fig:methods}
    \vspace{-0.5cm}
\end{figure}

Leveraging this state–action formulation, GPS-MTM employs a \textbf{bi-directional Transformer}~\cite{devlin2019bert} with a masked modeling objective, enabling contextually rich reconstruction of missing trajectory segments and the capture of semantic \emph{patterns of normalcy}~\cite{zhang_reespot_2024, gudavalli_reeframe_2024}, as illustrated in Figure~\ref{fig:methods}.

\subsection{Problem Formulation}

We represent a GPS trajectory $\mathcal{T}$ as a sequence of discrete \emph{stop-points}\\~\cite{ye_mining_2009, ying_semantic_2011}, each described by two concurrent modalities:  

\begin{enumerate}
   \item \textbf{POI category sequence} $\mathcal{S}_{POI} = (p_1, p_2, \ldots, p_N)$, where each $p_i$ denotes the semantic category of the visited point-of-interest.  
    \item \textbf{Stay-point details} $\mathcal{S}_{details} = (d_1, d_2, \ldots, d_N)$, where $d_i = (\text{id}_i, \text{st\_time}_i, \text{end\_time}_i, \text{st\_loc}_i)$ encodes temporal and spatial attributes of the stop.  
\end{enumerate}
\begin{equation*}
d_i = (\textit{id}_i, \textit{st\_time}_i, \textit{end\_time}_i, \textit{st\_loc}_i)
\end{equation*}

Given a trajectory $\mathcal{T}$, we partition it into an observed subset $\mathcal{T}_{obs}$ and a masked subset $\mathcal{T}_{mask}$. The task is to reconstruct $\mathcal{T}_{mask}$ conditioned on $\mathcal{T}_{obs}$, i.e., to approximate the conditional distribution $
P(\mathcal{T}_{mask} \mid \mathcal{T}_{obs})
$. The model parameters $\theta$ are estimated via Maximum Likelihood Estimation (MLE), a technique central to masked modeling approaches~\cite{devlin2019bert}:

\begin{equation}
\label{eq:mle}
\theta^* = \arg \max_{\theta} \prod_{p \in T_{mask}} \hat{P}_\theta(p \mid T_{obs})
\end{equation}

This formulation casts trajectory recovery as a probabilistic masked modeling problem~\cite{mtm}, enabling GPS-MTM to learn cross-modal dependencies between states and actions directly from raw trajectories, without manual labels.

\subsection{Model Architecture and Optimization}
Our GPS-MTM architecture consists of a multi-modal input embedding layer, a Transformer encoder, and task-specific prediction heads. Each stop-point, defined by a POI category $p_i$ and detail vector $d_i$, is mapped into a token embedding with modality-specific and temporal embeddings to preserve heterogeneity and order~\cite{radford2021clip}. The embedded sequence, including masked placeholders, is processed by a multi-layer Transformer encoder~\cite{devlin2019bert} that captures long-range trajectory dependencies via self-attention and feed-forward layers, yielding contextualized token representations. Masked-token embeddings are then passed to two heads: a \textbf{classification head} for POI prediction and a \textbf{regression head} for continuous detail reconstruction.

To train the network, we employ a composite loss function $\mathcal{L}$ that jointly optimizes both tasks, a common strategy in multi-task learning~\cite{deepam}. It combines the \textbf{Focal Loss} ($\mathcal{L}_{cls}$) for the classification task with the \textbf{Mean Squared Error (MSE)} ($\mathcal{L}_{reg}$) for the regression task. The final objective is to minimize the sum of these losses over all masked points $p_i \in T_{mask}$:
\begin{align}
\mathcal{L}_{cls} &= \sum_{p_i \in T_{mask}} -\alpha(1 - \hat{P}(p_i \mid T_{obs}))^\gamma \log(\hat{P}(p_i \mid T_{obs})) \\
\mathcal{L}_{reg} &= \sum_{p_i \in T_{mask}} \|d_i - \hat{d_i}\|_2^2 \\
\mathcal{L} &= \mathcal{L}_{cls} + \lambda \mathcal{L}_{reg} \label{eq:loss}
\end{align}
where $d_i$ and $\hat{d_i}$ are the ground-truth and predicted detail vectors. In the Focal Loss, $\alpha$ ($=0.5$) and $\gamma$ ($=2$) are the weighting and focusing parameters, respectively, and the hyperparameter $\lambda$ balances the contribution of the two loss components.

\section{Experiments}
\label{sec:experiments}

\definecolor{headercolor}{HTML}{F1F5F9} 
\definecolor{rowcolor}{HTML}{F8FAFC}    
\definecolor{rulecolor}{HTML}{D1D5DB}   

\begin{table*}[!t]
\centering
\sisetup{
  table-format=1.2,
  detect-weight=true,
  detect-inline-weight=math,
  tight-spacing=true
}

\setlength{\tabcolsep}{6pt}

{\arrayrulecolor{rulecolor}
\rowcolors{4}{rowcolor}{white}

\begin{tabular*}{\textwidth}{
  @{\extracolsep{\fill}}
  l l
  S S S
  S S S
  S S S
  S S S
  @{}
}
\toprule
& & \multicolumn{3}{c}{\textbf{ID}} & \multicolumn{3}{c}{\textbf{FD}} & \multicolumn{3}{c}{\textbf{Random}} & \multicolumn{3}{c}{\textbf{Goal}} \\
\cmidrule(lr){3-5} \cmidrule(lr){6-8} \cmidrule(lr){9-11} \cmidrule(lr){12-14}
\rowcolor{headercolor}
\multicolumn{2}{@{}l}{\multirow{-2}{*}{\textbf{Datasets}}} &
{Acc.} & {Rec.} & {Bias} &
{Acc.} & {Rec.} & {Bias} &
{Acc.} & {Rec.} & {Bias} &
{Acc.} & {Rec.} & {Bias} \\
\midrule

\multicolumn{2}{@{}l}{\textbf{Numosim-LA}} & 0.65 & 0.88 & 1.32 & 0.75 & 0.92 & 1.24 & 0.60 & 0.93 & 1.54 & 0.63 & 0.97 & 1.66 \\
\midrule

\multicolumn{14}{@{}l}{\textbf{UA-Atlanta}} \\
 & combined & 0.55 & 0.31 & 0.96 & 0.53 & 0.34 & 1.01 & 0.39 & 0.21 & 1.00 & 0.37 & 0.38 & 1.19 \\
 & hunger   & 0.38 & 0.21 & 0.91 & 0.37 & 0.22 & 0.86 & 0.35 & 0.19 & 0.93 & 0.28 & 0.05 & 0.79 \\
 & interest & 0.45 & 0.36 & 1.18 & 0.52 & 0.39 & 1.16 & 0.57 & 0.61 & 1.40 & 0.34 & 0.53 & 1.70 \\
 & social   & 0.43 & 0.39 & 1.34 & 0.49 & 0.32 & 1.16 & 0.56 & 0.35 & 1.19 & 0.34 & 0.26 & 0.90 \\
 & work     & 0.45 & 0.64 & 1.86 & 0.51 & 0.60 & 1.67 & 0.59 & 0.69 & 1.51 & 0.37 & 0.60 & 1.68 \\
\midrule

\multicolumn{14}{@{}l}{\textbf{UA-Berlin}} \\
 & combined & 0.51 & 0.40 & 1.02 & 0.46 & 0.48 & 1.05 & 0.47 & 0.19 & 0.91 & 0.43 & 0.29 & 0.93 \\
 & hunger   & 0.51 & 0.32 & 0.94 & 0.50 & 0.40 & 1.01 & 0.48 & 0.26 & 0.93 & 0.45 & 0.40 & 1.03 \\
 & interest & 0.47 & 0.25 & 0.97 & 0.52 & 0.27 & 1.16 & 0.57 & 0.46 & 1.11 & 0.40 & 0.54 & 1.06 \\
 & social   & 0.41 & 0.54 & 1.48 & 0.44 & 0.48 & 1.47 & 0.54 & 0.44 & 1.26 & 0.40 & 0.55 & 1.51 \\
 & work     & 0.53 & 0.42 & 1.02 & 0.54 & 0.54 & 1.10 & 0.50 & 0.33 & 0.98 & 0.48 & 0.47 & 1.04 \\
\midrule

\multicolumn{2}{@{}l}{\textbf{Geolife}} & 0.05 & 0.51 & 1.06 & 0.09 & 0.25 & 1.36 & 0.04 & 0.49 & 0.65 & 0.06 & 0.34 & 1.62 \\
\bottomrule
\end{tabular*}
}
\vspace{8pt}
\caption{Quantitative performance of GPS-MTM model on four downstream trajectory infilling tasks: Inverse Dynamics (ID), Forward Dynamics (FD), Random Masking, and Goal Prediction. Metrics: Accuracy (Acc.), Recall Range (Rec.), and Bias Ratio (Bias). Datasets: Numosim-LA (28 POI categories), UA-Atlanta and UA-Berlin (synthetic, 4 POI categories each), and Geolife (198 POI categories).}
\label{tab:main_results}
\end{table*}

In this section, we conduct a comprehensive evaluation of our proposed GPS-MTM model. We first describe the datasets used for training and evaluation. We then detail the experimental setup, including the baseline model and evaluation metrics. Finally, we present and analyze the quantitative results.

\subsection{Datasets}
We evaluate our model on four diverse trajectory datasets, each presenting unique challenges. \textbf{NUMOSIM} is a simulated dataset representing urban mobility patterns in Los Angeles with 28 unique Point-of-Interest (POI) categories. \textbf{Urban Anomalies} (UA) consists of synthetic GPS trajectories for two major cities, Atlanta (UA-Atlanta) and Berlin (UA-Berlin), each containing 4 primary POI categories; for fine-grained analysis, we further evaluate on five sub-tasks derived from user context: combined, hunger, interest, social, and work. Finally, \textbf{Geolife} is a large-scale real-world GPS trajectory dataset, from which we use a curated subset containing 198 unique POI categories, offering a challenging benchmark for scalability and performance on complex data.

\subsection{Experimental Setup}
\paragraph{Implementation Details.} Our GPS-MTM model consists of a \textbf{4-layer Transformer encoder} with a model dimension of \textbf{256}, \textbf{4 attention heads}, and a dropout rate of \textbf{0.1}. The model was implemented in \textbf{PyTorch}. For training, we used a batch size of \textbf{32} and optimized the model using the \textbf{AdamW optimizer} with a learning rate of $1 \times 10^{-4}$. The loss balancing hyperparameter $\lambda$ in Equation~\eqref{eq:loss} was set to \textbf{0.5}. All experiments were conducted on a single NVIDIA A100 GPU.

\paragraph{Tasks and Evaluation Metrics.}
We report results on 4 specific tasks, namely: \textbf{Forward Dynamics (FD)}, \textbf{Inverse Dynamics (ID)}, \textbf{Random masking}, and \textbf{GOAL}. As depicted in Figure~2, these tasks evaluate the model's ability to predict future movements (FD), infer past history (ID), handle realistic data gaps (Random), and predict final destinations (Goal). To provide a complete assessment, we evaluate performance using a comprehensive set of three metrics:

\begin{itemize}
    \item \textbf{Overall Accuracy}, defined as $\frac{1}{N} \sum_{i=1}^{N} \mathbb{I}(\hat{p}_i = p_i)$, measures the fundamental correctness of the predictions over all $N$ masked points.
    \item \textbf{Recall Range}, calculated as $\max_{k}(\text{Recall}_k) - \min_{k}(\text{Recall}_k)$ across all POI classes $k$, evaluates model consistency. A smaller range indicates more equitable performance across both common and rare location types.
   $\text{Bias Ratio} = \frac{P(\text{Predicted} = \text{Majority Class})}{P(\text{Actual} = \text{Majority Class})}$, assesses model fairness. A ratio closer to 1.0 signifies that the model's prediction distribution faithfully mirrors the ground-truth distribution.
\end{itemize}
The results for these metrics are shown in Table~\ref{tab:main_results}
.
\subsection{Results and Analysis}

The quantitative performance of GPS-MTM across all datasets and tasks is presented in Table~\ref{tab:main_results}. The results highlight the model's robust ability to learn complex mobility patterns, with performance nuanced by the nature and complexity of the underlying data.

On the simulated \textbf{Numosim-LA} dataset, the model establishes a strong baseline, achieving high accuracy (e.g., 0.75 on FD) and a low recall range, demonstrating proficiency in a controlled environment with a moderate number but unbalanced distribution of POI categories.

Performance on the synthetic \textbf{Urban Anomalies (UA)} datasets reveals the model's sensitivity to diverse behavioral contexts. For instance, on UA-Atlanta, GPS-MTM achieves accuracies of 0.56, 0.56, and 0.59, respectively, on the highly structured `interest', `social' and `work' sub-tasks, indicating its strength in capturing regular, predictable patterns. In contrast, performance in more varied tasks like `hunger' is lower. In particular, results consistently show low bias and minimal recall range, suggesting that the model identified a deterministic, low-entropy pattern in that specific data partition.

The most challenging task is the real-world \textbf{Geolife} dataset, with its large label space of 198 POI categories. As expected, raw accuracy is low (e.g., 0.05 for ID). However, a crucial finding is that the Bias Ratio remains close to 1.0 (e.g., 1.06 for ID). This demonstrates a key our formulation: despite the difficulty, it adapts to learn the true distribution of rare and common locations rather than collapsing to majority-class prediction.


\section{Future Work and Discussion}
\label{sec:discussion}

\noindent
Future research will focus on several key extensions. First, the model's learned representations of typical mobility can be leveraged for \textbf{anomaly detection}, flagging deviations from established \emph{patterns of normalcy} by comparing generated and observed sub-trajectories~\cite{zhang_reespot_2024, gudavalli_reeframe_2024, amiri_urban_2024, liu_online_2020, wen_uncertainty-aware_2025}. Second, these representations enable \textbf{controllable synthetic trajectory generation} under user-defined constraints, supporting applications in urban simulation~\cite{Krajzewicz2010, kim_humonet_2024} and data synthesis~\cite{stanford_numosim_2024, deepam, trajgpt}. Further work includes \textbf{multi-resolution temporal extensions} to generate continuous spatio-temporal paths and forecast future events~\cite{qin_transgte_2024}, enhancing \textbf{robustness to noisy GPS data} common in real-world deployments~\cite{tenzer_geospatial_2023}, and pursuing \textbf{semantic enrichment} by integrating geo-specialized language models like SpaBERT~\cite{li_spabert_2022} to infer POI categories automatically. These directions underscore the vision of treating mobility data as a \textbf{foundation-model modality}, enabling scalable representation learning for applications in urban analytics, security, and public health.

\vspace{-10pt}

\section{Acknowledgements}
This work is supported by the Intelligence Advanced Research Projects Activity (IARPA) via Department of Interior/ Interior Business Center (DOI/IBC) contract number 140D0423C0057. The U.S. Government is authorized to reproduce and distribute reprints for Governmental purposes notwithstanding any copyright annotation thereon. Disclaimer: The views and conclusions contained herein are those of the authors and should not be interpreted as necessarily representing the official policies or endorsements, either expressed or implied, of IARPA, DOI/IBC, or the U.S. Government. We would like to thank Kin Gwn Lore for insights and assistance during the initial phase of this project.


\bibliographystyle{plain}
\bibliography{main}

\end{document}